# Network with Sub-Networks

**Ninnart Fuengfusin**
*Graduate School of Life Science and Systems Engineering, Kyushu Institute of Technology,*
*2-4 Hibikino, Wakamatsu-ku, Kitakyushu, Fukuoka, 808-0196, Japan*
**Hakaru Tamukoh**
*Graduate School of Life Science and Systems Engineering, Kyushu Institute of Technology,*
*2-4 Hibikino, Wakamatsu-ku, Kitakyushu, Fukuoka, 808-0196, Japan*
*E-mail: fuengfusin-ninnart553@brain.kyutech.ac.jp, tamukoh@brain.kyutech.jp*
*http://www.lsse.kyutech.ac.jp/english/*

**Abstract**

We introduce *network with sub-networks*, a neural network which it's weight layers could be detached into sub-neural networks during inference. To develop weights and biases which could be inserted in both base and sub-neural networks, firstly, the parameters are copied from sub-model to base-model. Each model is forward-propagated separately. Gradients from a pair of networks are averaged and, used to update both networks. Our base model achieves the test-accuracy which is comparable to the regularly trained models, while the model maintains the ability to detach weight layers.

## 1. Introduction

Deep neural networks (DNNs) have been gained the attraction in the most recent years from their ability to provide the state-of-the-art performance in varied applications. However, to deploy those DNNs into the mobile devices is proved to problematic from the mobile devices are diverse in the specification. This raises the question: how to effectively design DNNs by given the specification of the mobile phone? To answer this question, two main factors within DNNs could be optimized.

The first factor is the performance of DNNs. In general, DNNs are provided an assumption by stacking the number of weight layers of DNNs, the better the performance of the model will be. One of the widely used example is the growing trend in the number of weight layers in ImageNet Large Scale Visual Recognition Competition (ILSVRC). AlexNet[1], the model which won ILSVRC-2012 consists of 8-weight layers. ResNet[2], the winner of ILSVRC-2015, contains of 152-weight layers. From AlexNet, ResNet reduces top-5 test error from 15.3 to 3.57. Even though, the growth in the number of weight layers might reduce the test-error rate of the model, it comes with the trade-off of the second factor, latency. More layers of DNNs means the higher number of parameters to compute. This also increases in the memory footprint which is crucial for the mobile device.

To solve this optimization problem, we might select the model which achieves the real-time performance given a mobile device specification. However, if the user conversely prefers the performance over the latency, this method does not satisfy the demand. Another method is to let the user select the preference and subsequently match the preference to the most suitable model. This method consumes the memory footprint for keeping various models into the mobile device is exceedingly huge. To satisfy user's preference in selectivity in both performance and latency without highly consuming memory footprint, we propose *network with sub-networks* (NSNs), DNNs which could be removed weight layers without dramatically decrease in the performance.

Generally, if one of the weight layers of DNNs is detached during the inference time, the performance of that model will dramatically be diminished. To explain our hypothesis, one of the widely used examples to explain how DNNs operate is to compare it as a feature extraction model. From the first weight layer, extracts the low-level features to the last layers extract the high level features. This creates a dependent relationship between each weight layer.

To challenge this concept, we propose the training method that allows NSNs to dynamically adapt to the removing of weight layers. We call this method, *copying learn-able parameters* and *sharing gradient*. Both methods are designed to optimize the learn-able parameters for both models, the model with or without the weight layer to detached.

## 2. Related Works

### 2.1. *Slimmable Neural Networks*

Slimmable Neural Networks[3] (SNNs) is the main inspired of this research. If our purposed method adds or remove weights in depth-wise direction, SNNs append or detach weights in width-wise direction. The range of possible width of networks requires to be pre-defined as the switch. The main research problem is the mean and variance of activations which come out from different-width weight layers are generally diverse. SNNs proposed *switchable batch normalization* to correct the mean and variance of SNNs.

## 3. Network with Sub-Networks

There are two types of models in NSNs: the base and sub-model. We define the base-model as DNNs with *n* hidden layers. Where *n* is a positive integer more than zero. From base-model, we could create n number of sub-models. Each of sub-model is mapped with *0,…,n-1* hidden layers. From this concept, the biggest sub-model takes all of the weight layers of the base-model except the input layer. The second biggest sub-model takes all of the weight layers of the biggest sub-model except the input layer of the biggest sub-model. This could be done repeatedly until we get the sub-model that has not any hidden layer.

In the next section, we will describe two of the processes in our purposed method: *copying learn-able parameters* and *sharing gradient*. Those processes are designed to be applied repeatedly in every mini-batch training.

### 3.1. *Copying Learn-able Parameters*

The goal of *copying learn-able parameters* is to combine each sub-model into the base-model. To enforce the similarity between weight and bias parameters between each model, the weights and biases are copied from the lesser sub-model to bigger sub-model and repeat until the base-model. The process is shown in Eq. (1) and Fig. 1.

Where $W_{o,m}$ is a weight variable, *o* is an integer indicating the order of weight layer and *m* is an integer indicating the model number.

$$W_{o+1,m+1} = W_{o,m} \qquad (1)$$

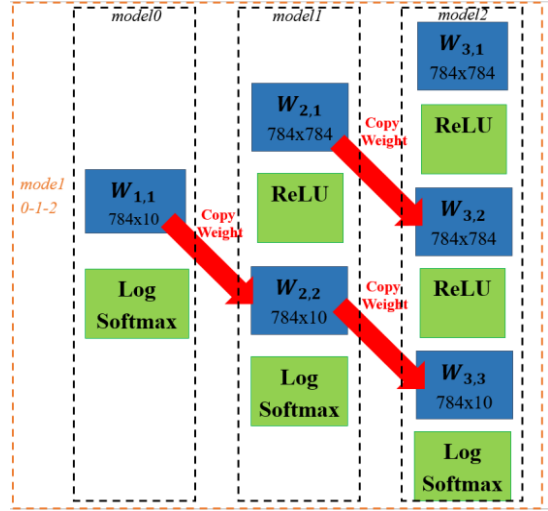

Fig. 1. Illustration of both *network with sub-networks* and *copying learn-able parameters* process. Where the base-model is two hidden layers DNNs and the sub-model as one hidden layer DNNs and a softmax-regression model. The name of the variable of weight, $W_{o,m}$ following with the size of weight array. Bias terms are excluded in this figure. *copying learn-able parameters* makes $W_{1,1}$, $W_{2,2}$ and $W_{3,3}$ to have exactly the same weight and bias variables..

After we apply this process, if we remove the input weight layer of base-model with the non-linear activation function, it will become the sub-model.

### 3.2. *Sharing Gradient*

*sharing gradient* is designed to constraint the learnable-variables to able to perform in two or more networks. Firstly, we forward propagate all of the models. During back propagation, the gradients from each model are collected separately. Each model is paired from the sub-model without the hidden layer to sub-model with a hidden layer until, the sub-model with *n-1* hidden layers to base-model. The gradients from each model's pair are averaged and updated the model's weights and bias. Overview of *sharing gradient* process is shown in Eq. (2) and Fig. 2 where *lr* is the learning rate and *L* is the loss function.

$$W_{m,o} = W_{m,o} - \frac{lr}{2}(\frac{\partial L_m}{\partial W_{m,o}} + \frac{\partial L_{m+1}}{\partial W_{m+1,o+1}}) \qquad (2)$$

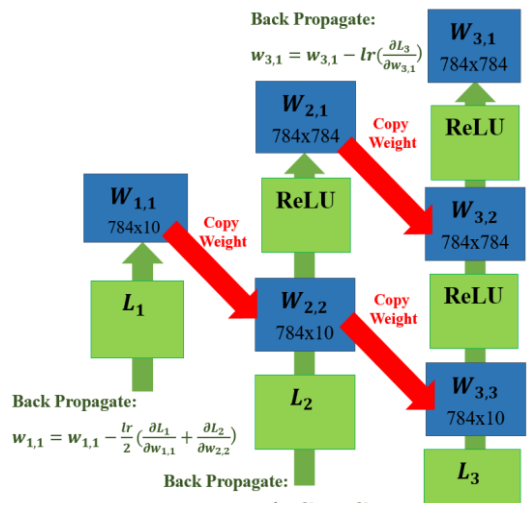

Fig. 2. *sharing gradient* in *model0-1-2* section. The gradients are shared from the sub-model to base-mode, pair by pair. Only the input weight layer of $W_{3,1}$ is regularly updated without sharing. Where $L_m$ is the loss function at the *m* model.

The reason behind *sharing gradient* for only a pair of models is from when sharing more than a pair gradient, the optimization becomes more complicated. In this case, the performance of NSNs hardly reaches the satisfiable point. Nonetheless, only an input layer of base-model, which has not a pair, is updated with the regular back propagation.

## 4. Experiments

The experiment was conducted using a hand-written digit image dataset, MNIST[4]. MNIST dataset consists of 60,000 training images and 10,000 test images. Each image in the dataset is the gray-scale image and, composed of 28x28 pixel. Each image pixel of MNIST mage was pre-processed into the range of [0, 1] by dividing all pixel value with 255.

Multi-layer perceptron (MLP) was applied with rectified linear unit (ReLU) as the non-linear activation function. The last layer was applied with log-softmax with the cost function as cross-entropy loss. The input layer of MLP was applied with Dropout[5], *p=0.8*. The hidden layers were put with dropout rate, *p=0.5*. In the case of softmax-regression, we did not apply dropout into the model since it was already under-fitting. The base-models were further regulated by using L2-weight penalty.

We applied stochastic gradient descent (SGD) with momentum, $\alpha = 0.9$. Although, we applied with slightly different format of SGD with momentum. From Tensorflow[6], neural networks framework, regular format of SGD with momentum was shown in Eq.3. Where *V* is the gradient accumulation term, *t* is the batch-wise iteration step and *G* is the gradient at *t+1*. Our format of SGD with momentum is shown in Eq. 4. After we found *V*, both of format was used the same Eq. 5. to update the weight, *W*.

$$V_{t+1} = \alpha V_t + G \qquad (3)$$
$$V_{t+1} = \alpha V_t + (1-\alpha)G \qquad (4)$$
$$W_{t+1} = W_{t+1} - lr(V_{t+1}) \qquad (5)$$

NSNs performed better with our format of SGD with momentum comparing the regular format at $\alpha = 0.9$. We speculated that NSNs required the higher proportion of the gradient accumulation, *V* comparing with the current gradient, *G* to converge. In the other hand, with the regularly trained DNNs, our format of SGD with momentum performed slightly worse in term of test accuracy. Hence, to perform a fair comparison between both type of models, the regularly-trained models were trained with Eq. 3. Our purposed method models were trained Eq. 4.

We set the training batch as 128. Each model had been training for 600 epoch. We reported the best test accuracy which might occur during the training. The initial learning rate, $lr = 0.3$ and step down by one third every 200 epoch.

The experimental result consists of two sections. First section is *model0-1* or the base-model as MLP with a hidden layer, *model1*, with a sub-model as the soft-max regression, *model0*. Second section is *model0-1-2* or the base-model as two layers MLP, *model2*. The sub-models are MLP with a hidden layer, *model1*, and the soft-max regression, *model0*. The graphical of *model0-1-2* is shown in Fig. 2. The base-line models which are regularly trained are referred as *ref-model* and following with number hidden layers. For example, *ref-model1* is the base-line MLP with a hidden layer. The results of base-line model are shown in Table. 1.

Table 1. Results of MNIST classification of base-line.

|  | Test Accuracy | Number Parameters | Regularization Parameter |
|---|---|---|---|
| *ref-model2* | 0.9886 | 1.24M | $1 \times 10^{-5}$ |
| *ref-model1* | 0.9882 | 0.62M | $5 \times 10^{-6}$ |
| *ref-model0* | 0.9241 | 7.85k | $9 \times 10^{-5}$ |

### 4.1. *Model0-1*

MLP with a hidden layer was used as the base-model. The sub-model was the softmax-regression. In all of the following experiment, we prioritized the base-model performance hence we reported all of the test accuracy of models in epoch that contains the best test accuracy of the base-model. *model0-1* results are displayed in Table 2. We applied the regularization parameter as $9 \times 10^{-6}$ at the base-model.

Table 2. Results of MNIST classification of *model0-1*.

|  | Test Accuracy | Number Parameters |
|---|---|---|
| *model1* | 0.9857 | 0.62M |
| *model0* | 0.9253 | 7.85k |

Comparing with the *ref-model1* and *model1*, the test accuracy of *model1* were dropped for an extent. This indicated that our purpose methods negatively affected the performance of the model for ability to removing the weight layers.

### 4.2. *Model0-1-2*

MLP with two hidden layer was used as the base-model. The sub-models were MLP with a hidden layer and the softmax-regression. *model0-1-2* results were displayed in Table 3. We applied the regularization parameter as $9 \times 10^{-5}$ at the base-model.

Table 3. Results of MNIST classification of *model0-1-2*.

|  | Test Accuracy | Number Parameters |
|---|---|---|
| *model2* | 0.989 | 1.24M |
| *model1* | 0.9843 | 0.62M |
| *model0* | 0.926 | 7.85k |

The difference in test accuracy between *model1* and *model2* indicated the bias of our purposed method towards the base-model. We speculated this bias might come from the *sharing gradient* process. All of the gradients that sub-models received, were averaged from multi-models. However, our base-model had an input layer that updated from gradient from the model itself as shown in Fig. 2.

Comparing with the result of *model1* in model*0-1* and *ref-model1*, our *model2* in *model0-1-2* contrastingly out-performed with *ref-model2* for a tiny margin. We hypothesized that the constraints of our purposed method might cause some type of regularization into the models. In case of *model1* in *model0-1*, this regularization effect might excessively strong and negatively affected the performance. Nevertheless, in case of *model2* in *model0-1-2*, the regularization effect seems to be adequate and positively affected the accuracy.

### 5. Conclusion

We purpose NSNs, DNNs that could be removed weight layers on fly. NSNs consists of a base-model and, sub-models. To assemble sub-models into the base-model, *copying learn-able parameters* is introduced. *sharing gradient* is applied for learn-able parameters could be used in two or more models. Our purposed method was conducted in the small scale experiment with a few hidden layers DNNs with MNIST dataset. The bigger-scale models with the dataset will be focused on future works.

**Acknowledgements**

This research was supported by JSPS KAKENHI Grant Numbers 17K20010.